\documentclass[sigconf]{acmart}
\AtBeginDocument{%
  \providecommand\BibTeX{{%
    \normalfont B\kern-0.5em{\scshape i\kern-0.25em b}\kern-0.8em\TeX}}}

\copyrightyear{2022}
\acmYear{2022}
\setcopyright{acmcopyright}\acmConference[MM '22]{Proceedings of the 30th ACM International Conference on Multimedia}{October 10--14, 2022}{Lisboa, Portugal}
\acmBooktitle{Proceedings of the 30th ACM International Conference on Multimedia (MM '22), October 10--14, 2022, Lisboa, Portugal}
\acmPrice{15.00}
\acmDOI{XXXXXXX.XXXXXXX}
\acmISBN{XXXXXX/XX/XX}

\acmSubmissionID{mmfp1744}

\usepackage{ulem}
\usepackage{comment}

\usepackage{booktabs}
\usepackage{multirow}
\usepackage{subfigure}

\usepackage{algorithm}
\usepackage{algorithmic}

\usepackage{balance}
\usepackage{bbding}

\def\eg{{\itshape e.g.}}
\def\ie{{\itshape i.e.}}
\def\etc{{\itshape etc.}}
\def\wrt{{\itshape w.r.t.}}

\begin{document}

\title{MimCo: Masked Image Modeling Pre-training with Contrastive Teacher
}

\iftrue
    \author{Qiang Zhou}
    \affiliation{
        \institution{Alibaba Group}
        \city{Hangzhou}
        \country{China}
    }
    \email{jianchong.zq@alibaba-inc.com}
    
    \author{Chaohui Yu}
    \affiliation{
        \institution{Alibaba Group}
        \city{Hangzhou}
        \country{China}
    }
    \email{huakun.ych@alibaba-inc.com}
    
    \author{Hao Luo}
    \affiliation{
        \institution{Alibaba Group}
        \city{Hangzhou}
        \country{China}
    }
    \email{michuan.lh@alibaba-inc.com}
    
    \author{Zhibin Wang\textsuperscript{\Envelope}}
    % \authornote{Corresponding author}
    \thanks{\Envelope~Corresponding author}
    \affiliation{
        \institution{Alibaba Group}
        \city{Hangzhou}
        \country{China}
    }
    \email{zhibin.waz@alibaba-inc.com}
    
    \author{Hao Li}
    \affiliation{
        \institution{Alibaba Group}
        \city{Hangzhou}
        \country{China}
    }
    \email{lihao.lh@alibaba-inc.com}
\fi

\renewcommand{\shortauthors}{Qiang Zhou et al.}

\begin{abstract}
Recent masked image modeling (MIM) has received much attention in self-supervised learning (SSL), which requires the target model to recover the masked part of the input image.
Although MIM-based pre-training methods achieve new state-of-the-art performance when transferred to many downstream tasks, the visualizations show that the learned representations are less separable, especially compared to those based on contrastive learning pre-training.
This inspires us to think whether the linear separability of MIM pre-trained representation can be further improved, thereby improving the pre-training performance.
Since MIM and contrastive learning tend to utilize different data augmentations and training strategies, combining these two pretext tasks is not trivial.
In this work, we propose a novel and flexible pre-training framework, named MimCo, which combines MIM and contrastive learning through two-stage pre-training.
Specifically, MimCo takes a 
pre-trained contrastive learning
model as the teacher model and is pre-trained with two types of learning targets: patch-level and image-level reconstruction losses.

Extensive transfer experiments on downstream tasks demonstrate the superior performance of our MimCo pre-training framework.
Taking ViT-S as an example, when using the pre-trained MoCov3-ViT-S as the teacher model, MimCo only needs 100 epochs of pre-training to achieve 82.53\% top-1 finetuning accuracy on Imagenet-1K, which outperforms the state-of-the-art self-supervised learning counterparts.

\end{abstract}




\begin{teaserfigure}
  \centering 
  \includegraphics[width=1.0\textwidth]{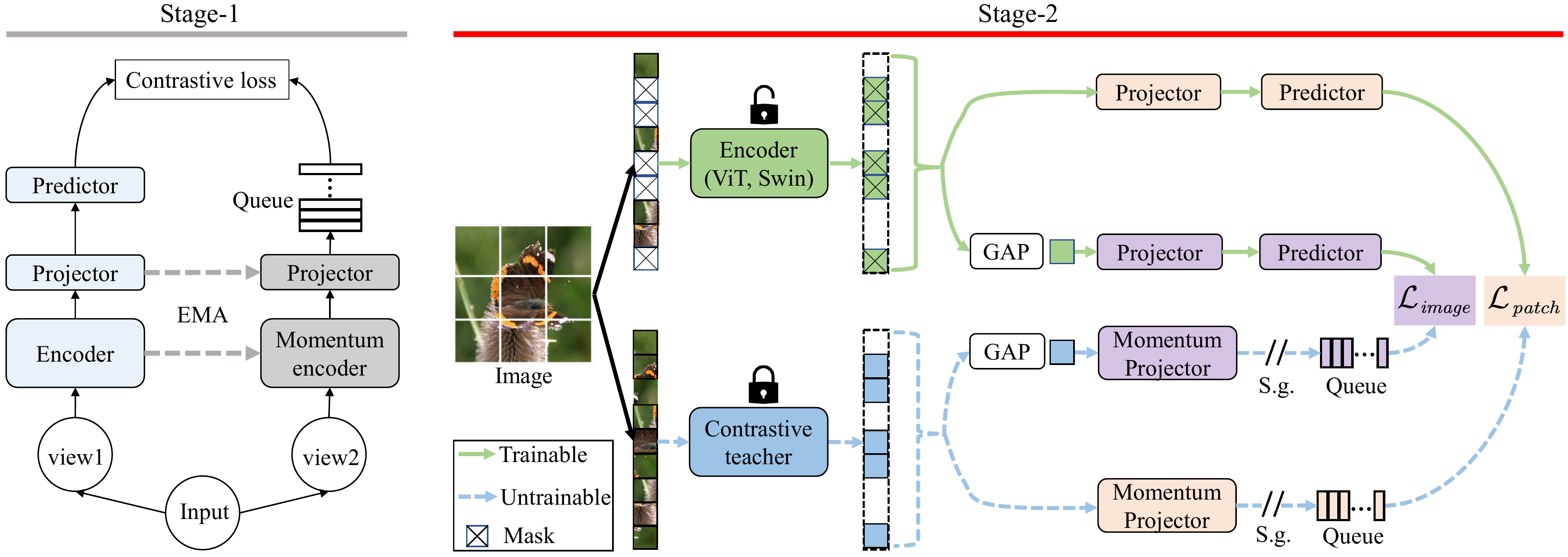}
  \caption{The proposed pre-training framework MimCo. MimCo is pre-trained in two stages. The first stage (Stage-1) denotes contrastive learning-based pre-training, \eg, MoCov3~\cite{chen2021mocov3}, MoBY~\cite{xie2021swinselfsupervised}. The second stage (Stage-2) is our main framework, which uses the pre-trained encoder in stage-1 as a contrastive teacher. ``S.g.'' denotes stop gradient, ``GAP'' is a global average pooling layer.}
  \Description{The left part depicts the architecture of stage-1 pre-training, \eg, MoCov3 and MoBY. The right part depicts our MimCo framework, which has two branches. The first branch takes as input masked image while the other tasks as input unmasked image using the encoder (pre-trained in stage-1) as a contrastive teacher. Then, two types of reconstruction losses (patch-level and image-level) are computed between the features of these two branches.}
  \label{fig:arch}
\end{teaserfigure}

\maketitle

\section{Introduction}

% \begin{figure}[h]
% \centering  
% \subfigure[T-SNE feature visualization]{   
% \begin{minipage}{6cm}
%     \includegraphics[width=\linewidth]{figures/motivation.pdf} 
% \end{minipage} 
% }
% \subfigure[Training efficiency comparison]{  
% \begin{minipage}{6cm}
% \centering    
%     \includegraphics[width=\linewidth]{figures/motivation2.pdf}  
% \end{minipage}
% }
% \caption{
% (a)~T-SNE feature visualization of MIM method SimMIM~\cite{xie2021simmim}, contrastive learning method MoBY~\cite{xie2021swinselfsupervised}, and our MimCo on ImageNet-1K validation dataset.
% %
% (b)~Finetune accuracy on ImageNet-1K \wrt~effective pre-training epochs based on ViT-S/16 architecture.
% %
% Our MimCo exhibits both better feature discrimination and higher pre-training efficiency.} 
% \label{fig:motivation}   
% \end{figure}

%``The revolution will not be supervised.'' --- LeCun. 
%
With the development of deep neural networks~\cite{he2016deep} and transformers~\cite{vaswani2017attention}, masked language modeling (MLM) has achieved great success and emerged as an important  self-supervised pre-training approach for language models in natural language processing (NLP).
For instance, BERT~\cite{devlin2018bert} innovatively proposes to randomly mask a part of the input sequence and learn to predict or reconstruct these masked tokens, which has almost become the standard pre-training paradigm in NLP.
Inspired by the success of MLM, recently, masked image modeling (MIM) has achieved fast development in visual pre-training tasks, showing the potential to be an important training paradigm for self-supervised learning in vision. 

MIM is a task of randomly masking some patches of an input image and learning to reconstruct the masked patches. ViT~\cite{dosovitskiy2020vits} and BEiT~\cite{bao2021beit} propose to perform MIM in self-supervised pre-training with vision transformer (ViT)~\cite{dosovitskiy2020vits}. BEiT first proposes to use a trained discrete variational autoencoder (dVAE)~\cite{ramesh2021zero} to build a visual vocabulary, imitating the language vocabulary in NLP, which provides promising performance in visual pre-training.
Following BEiT, very recently, several MIM literature have been proposed to further promote the self-supervised learning in vision. Some methods~\cite{he2021mae,xie2021simmim} propose to directly regress the raw pixels of the masked patches in a simple and effective way. Other methods~\cite{wei2021maskedfeat,zhou2021ibot,dong2021peco} turn to improve the semantic of visual tokens.

Although state-of-the-art MIM-based self-supervised learning methods achieve impressive performance when transferred to downstream tasks, they suffer from 
%\sout{\textbf{slow pre-training} and} 
\textbf{poor linear separability of learned representations}, as shown in Figure~\ref{fig:motivation}. 
%\textcolor{red}{As shown in Figure~\ref{fig:arch}(b), we randomly choose 10 classes of ImageNet-1K validation set to visualize for simplicity, SimMIM~\cite{xie2021simmim} is not good at learning linear separable features.}
%
%\sout{For example, the MAE~\cite{he2021mae} models need to be pre-trained for 1600 epochs to achieve satisfactory performance. This is quite expensive when pre-training large models with more data.}
%
The linear separability of representations is highly correlated with transfer performance for tasks that require frozen features, such as image retrieval.
%Compared with previous contrastive learning based pre-training methods, recent MIM-based methods perform poorly on these tasks.
It is not surprising that recent MIM-based pre-taining work~\cite{xie2021simmim} has limited performance on these downstream tasks.
In contrast, the learned representations are more linearly separable based on the contrastive learning pre-training paradigm, \eg, MobY~\cite{xie2021swinselfsupervised} and MoCov3~\cite{chen2021mocov3}. 
%\sout{In this work, we develop a novel MIM pre-training framework, named MimCo, to address these two problems.}
%
This motivates us to combine these two pre-training paradigms of MIM and contrastive learning and propose a new pre-training framework. %The architecture of MimCo is shown in Figure~\ref{fig:arch}(a).

%First, MimCo is flexible by decoupling the MIM and contrastive learning paradigms.
%As shown in Figure xx, the contrastive learning pretrained model is used as the contrastive teacher model, similar to role of discrete VAE (dVAE)~\cite{ramesh2021zero} in BEiT~\cite{bao2021beit}.
%Flexibility is mainly reflected in two aspects. On the one hand, MIM and contrastive learning are two different pre-training paradigms, differing vastly from data augmentation to training hyperparameters, and being pretrained respectively is more convenient and flexible.

However, introducing contrastive learning into MIM is not trivial since MIM and contrastive learning
tend to utilize different data augmentations and training strategies.
In this work, we propose a novel and flexible pre-training framework, named MimCo. As show in Figure~\ref{fig:arch},  MimCo is pre-trained in two stages.
%instead of combining MIM and contrastive learning through naive multi-task learning, our MimCo is pre-trained in two stages.
In the first stage, the contrastive teacher model is pre-trained based on contrastive learning methods, such as MoCov3~\cite{chen2021mocov3}, MoBY~\cite{xie2021swinselfsupervised}, \etc.
In the second stage, MimCo is pre-trained with MIM, and the contrastive teacher model will not be updated, which is similar to the role of dVAE~\cite{ramesh2021zero} in BEiT~\cite{bao2021beit}.
Through decoupling the MIM and contrastive learning paradigms, MimCo is more flexible and efficiency during pre-training. 
First, MIM and contrastive learning are two different pre-training paradigms, differing vastly from data augmentations to training hyperparameters
%\textcolor{blue}{, and being pre-trained individually is more convenient and flexible.}
. Thus pre-training them individually is more convenient and flexible.
Second, advances in contrastive learning based pre-training will benefit MimCo by simply replacing the contrastive teacher with a new and better model.

%Unlike MIM pre-training frameworks such as BEiT~\cite{bao2021beit}, which first learn image tokenizer via discrete VAE (dVAE)~\cite{ramesh2021zero}, our pre-training framework directly uses a pre-trained contrastive learning based model as the image tokenizer.
%For simplicity, we call this image tokenizer the contrastive teacher model.
%With the assistance of the contrastive teacher model, our pre-training framework is quite efficient and the features are highly linearly separable.
%For example, as shown in Table~\ref{tbl:main_cls}, when using Swin-T as the backbone and pre-training for only 100 epochs, MimCo achieves a finetuning accuracy of 81.64\% on ImageNet-1K dataset, outperforming existing methods.
To take full advantage of the contrastive teacher model, we further propose two types of reconstruction losses.
The first is the \textbf{patch-level} reconstruction loss. For the masked patches, we take the corresponding features of the contrastive teacher model as reconstruction targets.
Compared to directly predicting the patch features~\cite{wei2021maskedfeat}, we propose to reconstruct the patch features through a contrastive loss, which performs better.
The second is the \textbf{image-level} reconstruction loss, which reconstructs the overall features of the masked image.
The image-level reconstruction, also implemented as a contrastive loss, helps improve the linear separability of learned representations, as shown in Table~\ref{tbl:ablation_loss_retrieval}.

Overall, this work makes the following contributions:
\begin{itemize}
    \item We propose a novel and flexible pre-training framework, named MimCo, which takes a contrastive learning pre-trained model as the teacher model. 
    Compared with recent MIM pre-training methods, MimCo owns more separable representations and better transfer performances.
    
    \item To take full advantage of the contrastive teacher model, we propose two reconstruction losses, \ie, patch-level and image-level, which are experimentally verified to be effective.
    
    %\item We reveal the inadequacies of MIM pretraining methods through feature visualization and experiments. For example, the finetune accuracy of MIM is much worse than contrastive learning when the amount of labeled data is small.
    
    %\item We propose a more flexible and efficient pretraining framework to combine MIM and CL, named MimCo, outperforming the naive solution based on multi-task learning.
    
    %\item \textcolor{blue}{Current MIM methods usually need longer training time, we propose to extract and leverage the knowledge from models pretrained via contrastive learning to maintain training flexibility and efficiency.}
    
    \item Extensive experiments on many downstream tasks, including classification, object detection, instance segmentation, and semantic segmentation demonstrate that our MimCo pre-training framework can achieve superior transfer performance against state-of-the-art methods.
    
\end{itemize}

\section{Related Work}
During the booming of deep learning, recent years have witnessed remarkable progress of self-supervised learning (SSL)~\cite{doersch2015unsupervised,wang2015unsupervised,noroozi2016unsupervised,zhang2016colorful,pathak2016context,pathak2017learning,gidaris2018unsupervised}. 

\paragraph{\textbf{Contrastive Learning Pre-training}}
%\sout{To learn general visual representations with unlabeled data, previous self-supervised learning methods often leverage different pretext tasks via a destroy and restore way, e.g.,  context prediction~\cite{doersch2015unsupervised},  jigsaw puzzles solving~\cite{noroozi2016unsupervised}, image colorization~\cite{zhang2016colorful}, image inpainting~\cite{pathak2016context}, rotation prediction~\cite{gidaris2018unsupervised}, et~al.}
%
Recently, one line of research focus on contrastive learning~\cite{alexey2016discriminative} based pre-training methods, and plenty of literature~\cite{dosovitskiy2014discriminative,van2018representation,wu2018unsupervised,cao2020parametric,chen2020simple,grill2020bootstrap,he2020momentum,xie2021propagate,xie2021swinselfsupervised} have been proposed, which dominate the previous self-supervised visual representation learning.
These methods learn discriminative representation by attracting similar instances and dispelling dissimilar instances, based on two or multiple different augmented views of one image. 
For instance, SimCLR~\cite{chen2020simple} proposes a simple framework to promote the performance of self-supervised learning by maximizing the mutual information between two augmented views of a image.
MoCo~\cite{he2020momentum} uses a momentum encoder to maintain consistent representations of negative pairs drawn from a memory bank, which enables building a large and consistent dictionary on-the-fly that facilitates contrastive unsupervised learning.
BYOL~\cite{grill2020bootstrap}  proposes a metric-learning manner, which uses a moving average network to produce prediction targets as a means of stabilizing the bootstrap step. 
MoBY~\cite{xie2021swinselfsupervised} proposes an elegant combination of MoCo~\cite{he2020momentum} and BYOL~\cite{grill2020bootstrap}, with a proper training recipe and lighter tricks, MoBY can achieve high performance. 
%
%Despide the success, contrastive learning based methods usually depend on strong data augmentations, which may limit the potential of self-supervised visual learning.
%
%\sout{Our method is complementary to these contrastive learning based methods by distilling the learned discriminative features.}
Our method takes the contrastive learning pre-trained model as the teacher model and aims to improve the performance of MIM pre-training.

\paragraph{\textbf{Masked Image Modeling Pre-training}}
Masked language modeling (MLM) methods~\cite{devlin2018bert,radford2018improving} often mask some part of the input sequence and then train the models to model the missing portion. MLM methods have been a popular language model pre-training paradigm in NLP.
Inspired by the great success of modern MLM methods in NLP,  very recently, another line of research on self-supervised visual learning tends to masked image modeling (MIM). 
iGPT~\cite{chen2020generative} trains a sequence Transformer~\cite{vaswani2017attention} to auto-regressively predict the next pixels and learns state-of-the-art representations for low resolution datasets.
ViT~\cite{dosovitskiy2020vits} proposes to predicts mean color of each corrupted patch using their respective patch representations with ViT. 
BEiT~\cite{bao2021beit} proposes to use a pre-trained discrete variational autoencoder (dVAE)~\cite{ramesh2021zero}, which can be seen as a offline tokenizer, to encode masked patches.
Following BEiT~\cite{bao2021beit}, MAE~\cite{he2021mae} develops an asymmetric encoder-decoder architecture to reconstruct the normalized masked patches.
SimMIM~\cite{xie2021simmim} propose a simple framework to reconstruct the raw pixels.
iBOT~\cite{zhou2021ibot}  performs masked image modeling via self-distillation by introducing an online tokenizer.
PeCo~\cite{dong2021peco} proposes to learn a perceptual codebook, which exhibits better semantic meanings of the visual tokens.
MaskFeat~\cite{wei2021maskedfeat} presents masked feature prediction with HOG~\cite{dalal2005histograms} for self-supervised pre-training of video models. 
Our method is complementary to the MIM methods.

\paragraph{\textbf{Self-supervised Learning and Knowledge Distillation}}
Knowledge distillation (KD)~\cite{ba2014deep,hinton2015distilling} aims to distill knowledge from a well-trained model (teacher) to another model (student). Typical KD methods usually leverage the intermediate features or the output logits of a teacher model to supervise the training of a student model. 
Hinton et~al.~\cite{hinton2015distilling} first proposes to distill knowledge from teacher's output logits into smaller student model.
FitNets~\cite{romero2014fitnets} extend this idea to distill the knowledge via minimizing the intermediate features learned by the teacher and the student model.
Recently, some works introduce the KD methods into self-supervised learning~\cite{noroozi2018boosting,tian2019contrastive,chen2020big,CompRess_KoohpayeganiTP20,SEED_FangWWZYL21}.
%
%NoisyStudent~\cite{xie2020self} proposes to distill the knowledge from soft pseudo-labels to unlabelled data in a self-training way.
%
\cite{noroozi2018boosting} proposes a knowledge transfer method to decouple the pre-training model and the final task model based on clustering the learned features.
\cite{tian2019contrastive} proposes to use contrastive loss to learn cross-modality consistency.
CompRess~\cite{CompRess_KoohpayeganiTP20} compresses an already learned deep self-supervised teacher model into a smaller student model by mimicking the relative similarity of data points in the teacher's embedding space.
SEED~\cite{SEED_FangWWZYL21} first trains a large network in a self-supervised fashion, and then trains a small network to mimic the similarity score distribution inferred by the large network over a set of instances.
DINO~\cite{caron2021dino} proposes to simplify self-supervised training by directly predicting the output of a teacher network, which is built with a momentum encoder.
%
%In this work, we propose an efficient knowledge distillation framework to jointly distill the pre-trained discriminative features of the contrastive learning based literature and optimize the MIM task.
In this work, we propose to extract knowledge from pre-trained contrastive teacher models when performing MIM pre-training.

\section{Approach}

We inspire our method by improving the performance of MIM pre-training with the assistance of contrastive learning.
%
%\textcolor{red}{Instead of using inefficient multi-task learning, we propose a novel and flexible pre-training framework.}
Instead of combining MIM and contrastive learning via multi-task learning, we propose a novel two-stage pre-training framework that is more flexible and achieves higher performance.
In this section, we elaborate the framework, learning targets, and implementation details of MimCo, respectively.

\subsection{Framework}
\label{sec:framework}
%MIM and contrastive learning are two of the most successful pretext tasks in self-supervised learning in computer vision.

MimCo is pre-trained in two-stages.
In the first stage, we use contrastive learning methods, such as MoCov3~\cite{chen2021mocov3}, MOBY~\cite{xie2021swinselfsupervised}, etc., to pre-train on the ImageNet-1K dataset.
The pre-trained model will be used as the contrastive teacher model in our MimCo pre-training, as shown in Figure~\ref{fig:arch}.
We refer readers to these works for more details, and in our experiments, we directly use the open-source models from these works.
%Below, we detail the framework and learning targets of the second pre-training stage of MimCo.

As shown in Figure~\ref{fig:arch}, MimCo mainly consists of a learnable encoder $\mathbf{f}$, a frozen contrastive teacher model $\mathbf{f^{'}}$, and two sets of contrastive learning modules. 
%The teacher model is pre-trained in advance using the contrastive learning pretext task. In this work, we use the pre-trained models from MoCov3~\cite{chen2021mocov3} and MoBY~\cite{xie2021swinselfsupervised}, two representative contrastive learning based SSL works.
%
During pre-training, for each training sample $x$, we first randomly generate a mask $m$ using the same masking strategy as in SimMIM~\cite{xie2021simmim}.
Then, the contrastive teacher model takes as input the non-masked image and extracts features $\mathbf{f^{'}}(x)$,
%extracts features $\mathbf{f^{'}}(x)$ for the non-masked image, 
while the learnable encoder extracts features $\textbf{f}(x, m)$ for the masked image using the generated mask $m$.
The non-masked features $\mathbf{f^{'}}(x)$ will be used as the targets to reconstruct the masked feature $\textbf{f}(x, m)$ through patch-level and image-level reconstruction losses, which will be described in the next section.
After pre-training, only the learnable encoder is applied to non-masked images to extract representations for downstream tasks.

%The frozen encoder is followed by a momentum projector (2-layer MLP), while the learnable encoder is followed by a MIM decoder (1-layer Convolution), and a projector (2-layer MLP) and a predictor (2-layer MLP), respectively.

%The MIM decoder, denoted as $\mathbf{g}$, aims to reconstruct knowledge using the feature of the frozen encoder as tokenizer, through a patch-level reconstruction loss and will be described in the next section.
%
%The patch-level reconstruction loss mainly focuses on the reconstruction of the local patches.
%
%In addition to reconstructing the local patch, we further propose to perform global view reconstruction through an image-level reconstruction loss, which is implemented as a contrastive loss between the features $\mathbf{f}(x, m)$ of the learnable encoder and the features $\mathbf{f}^{'}(x)$ of the frozen encoder.
%The projector (momentum projector) and predictor are used to extract global view features from the local dense features $\mathbf{f}(x, m)$ and $\mathbf{f}^{'}(x)$.

\subsection{Learning Targets}
\label{sec:learn_target}

In this section, we elaborate the learning targets of MimCo, including the 
%\textcolor{blue}{patch-level reconstruction loss and the image-level reconstruction loss.}
patch-level and image-level reconstruction losses.
Algorithm~\ref{alg:pseudo_code} provides the pseudo-code of MimCo for these learning targets.

\paragraph{\textbf{Patch-level Reconstruction Loss}}

Similar to other MIM-based SSL work~\cite{he2021mae,xie2021simmim}, we reconstruct knowledge for those masked patches of input sample $x$.
%We experimentally find that reconstructing the features of the frozen encoder outperforms directly reconstructing raw pixel colors.
MaskFeat~\cite{wei2021maskedfeat} verifies that reconstructing the features of the a pre-trained model via $\ell_1$-loss is better than directly reconstructing the raw pixels or HOG features.
Unlike MaskFeat, we experimentally find that reconstructing the features via contrastive loss is superior to $\ell_1$-loss, as shown in Table~\ref{tbl:loss_patch}.
%
%We experimentally find that reconstructing the features of the teacher model via contrastive learning outperforms reconstructing the features~\cite{wei2021maskedfeat} or raw pixels~\cite{xie2021simmim,he2021mae} via $\ell_1$-loss.
%
To be specific, we adopt a contrastive learning loss to model the similarity of the local patches between masked and non-masked images.
Following MoBY~\cite{xie2021swinselfsupervised}, a projector $\mathbf{p}_1^p$ (2 layer convolution), a predictor $\mathbf{p}_2^p$ (2 layer convolution), and a momentum projector $\mathbf{p}_3^p$ (2 layer convolution) are introduced when computing the contrastive loss, as shown in Figure~\ref{fig:arch}.
Formally, the patch-level reconstruction loss $\mathcal{L}_{\text{patch}}$ can be computed as follows. For convenience, we show the $\mathcal{L}_{\text{patch}}$ computed on one input sample $x$.
%
% \begin{equation}
%     L_{\text{intra}} = \frac{\sum \left [ \left |  \right | \mathbf{g}(\mathbf{f}(x)) - \mathbf{f}^{'}(x, m) \left |  \right |_{1} \cdot m \right ] }{\sum m} 
% \label{eq:intra_loss}
% \end{equation}
%
% \begin{equation}
%     \mathcal{L}_{\text{patch}} = \frac{1}{N} \sum^{N}_{i=1} \left [ \frac{\sum_{j} \left ( m^j_i \cdot \left |  \right | \mathbf{g}(\mathbf{f}(x_i, m_i)) - \mathbf{f}^{'}(x_i) \left |  \right |^{j}_{1} \right ) }{\sum_{j} m^j_i} \right ],
% \label{eq:intra_loss}
% \end{equation}
%
%
\begin{equation}
\mathcal{L}_{\text{patch}} = \frac{1}{M} \sum^{M}_{i=1} -\text{log}\frac{\text{exp}(q_i \cdot k_{(i,+)}/\tau)}{\text{exp}(q_i \cdot k_{(i,+)}/\tau) + \sum_{j=1}^{K} \text{exp}(q_j \cdot k_j /\tau)},
\label{eq:patch_loss}
\end{equation}
in which:
\begin{equation}
\left \{
    \begin{array}{ll}
        q_i = \mathbf{p}_2^p(\mathbf{p}_1^p(\mathbf{f}(x, m)))_i, \\
        k_{i,+} = \mathbf{p}_3^p(\mathbf{f}^{'}(x))_i, \\
    \end{array}
\right.
\end{equation}
where $M$ denotes the total number of masked patches of a sample $x$. $m \in \mathbb{R}^{1 \times \frac{H}{P} \times \frac{W}{P}}$ is the randomly generated mask applied to $x$. $\{\mathbf{p}_2^p(\mathbf{p}_1^p(\mathbf{f}(x, m))), \mathbf{p}_3^p(\mathbf{f}^{'}(x))\} \in \mathbb{R}^{C \times \frac{H}{P} \times \frac{W}{P}}$ are the output features of the predictor $\mathbf{p}_2^p$ and momentum projector $\mathbf{p}_3^p$, respectively.  $P$ denotes the patch size in ViTs and should take the stride value into consideration in Swins, which has downsampling operations.
$q_i, k_{i,+}$ are the feature vectors corresponding to the $i_{th}$ masked patch from the learnable encoder and frozen teacher model, respectively. $k_j$ is the $j_{th}$ feature vector in the $key$ queue. $K$ is the length of the $key$ queue (4096 by default). $\tau$ is a temperature term (0.2 by default).
Since patch features are very redundant, for image $x$, we instead put the average feature of all patch features of the teacher model into the key queue.

\paragraph{\textbf{Image-level Reconstruction Loss}}

As compensation for the patch-level reconstruction loss, which only focuses on local patch reconstruction, the image-level reconstruction loss here focuses on reconstruction from the global view.
We adopt a contrastive loss to encourage the global features between masked and non-masked images to be as similar as possible.
%so as to achieve the goal of reconstructing global features.
The difference from other contrastive learning-based SSL works~\cite{chen2021mocov3,xie2021swinselfsupervised} is that instead of taking two views of a sample as a positive pair, we take the non-masked view $x$ and the masked view $(x, m)$ as a positive pair.
For convenience, we denote the projector, predictor, and momentum projector as $\mathbf{p}^I_1$, $\mathbf{p}^I_2$, and $\mathbf{p}^I_3$, respectively, which are all 2 layer MLP. The image-level reconstruction loss $\mathcal{L}_\text{image}$ is computed as:
%
% \begin{equation}
%     \begin{split}
%         &q = \mathbf{p}_2(\mathbf{p}_1(\mathbf{f}(x, m))) \\
%         &k_+ = \mathbf{p}_3 (\mathbf{f}^{'}(x)) \\
%         &L_{\text{inter}} = -\text{log}\frac{\text{exp}(q\cdot k_{+}/\tau)}{\text{exp}(q\cdot k_{+}/\tau) + \sum_{i=1}^{K} \text{exp}(q\cdot k_i/\tau)} 
%     \end{split}
% \label{eq:inter_loss}
% \end{equation}
%
\begin{equation}
\mathcal{L}_{\text{image}} = -\text{log}\frac{\text{exp}(q\cdot k_{+}/\tau)}{\text{exp}(q\cdot k_{+}/\tau) + \sum_{i=1}^{K} \text{exp}(q\cdot k_i/\tau)},
\label{eq:inter_loss}
\end{equation}
in which:
\begin{equation}
\left \{
    \begin{array}{ll}
        q = \mathbf{p}^I_2(\mathbf{p}^I_1(\mathbf{f}(x, m))), \\
        k_+ = \mathbf{p}^I_3 (\mathbf{f}^{'}(x)), \\
    \end{array}
\right.
\end{equation}
where $q, k_+, k_i$ are all 1-D feature vectors. $k_i$ is the feature of unmasked images in the $key$ queue. $K$ is the length of the $key$ queue (4096 by default). $\tau$ is a temperature term (0.2 by default).

\begin{table*}
\caption{Finetuning accuracy on ImageNet-1K. {\itshape Sup.} denotes the supervised baselines. $^{\dagger}$ denotes using multi-crop augmentation.
$^{\ddag}$ denotes our pre-training results using official code.}
\label{tbl:main_cls}
\centering
\resizebox{0.73\linewidth}{!}{
\begin{tabular}{@{}l|c|c|c|c|l@{}}
\toprule
Method & Arch.                     & Extra model & Pre-train Epochs & Effective Epochs & Top-1 acc (\%) \\ \midrule
Sup.~\cite{liu2021swins}   & \multirow{5}{*}{Swin-T}   &            &        &                  &  \hspace{0.5cm}81.2            \\ 
SimMIM~\cite{xie2021simmim}   &    &            &  800      &        800                        &  \hspace{0.5cm}80.9$^\ddag$          \\ 
MoBY~\cite{xie2021swinselfsupervised}  &    &            &  300      &  600                     &  \hspace{0.5cm}81.4          \\ 
MimCo (Ours)  &                           & MoBY-Swin-T-300e           &  100      &       700         &  \hspace{0.5cm}81.7       \\
MimCo (Ours)  &                           & MoBY-Swin-T-300e           &  300      &          900      &  \hspace{0.5cm}\textbf{81.9}        \\
\midrule

Sup.~\cite{liu2021swins} &  \multirow{6}{*}{Swin-B}  & & &      & \hspace{0.5cm}83.5 \\
SimMIM~\cite{xie2021simmim} & &  & 100 & 100                    & \hspace{0.5cm}83.5 \\
SimMIM~\cite{xie2021simmim} & &  & 800 & 800                    & \hspace{0.5cm}84.0 \\
MoBY~\cite{xie2021swinselfsupervised} & & & 300 & 600           & \hspace{0.5cm}83.1$^\ddag$ \\
MimCo (Ours) & & MoBY-Swin-B-300e & 100 & 700                   & \hspace{0.5cm}84.0 \\
MimCo (Ours) & & MoBY-Swin-B-300e & 300 & 900                   & \hspace{0.5cm}\textbf{84.3} \\
\midrule

Sup.~\cite{touvron21deit}   & \multirow{8}{*}{ViT-S/16} &            &        &                  &  \hspace{0.5cm}79.9           \\ 
BEiT~\cite{bao2021beit}  &            &    dVAE        &    800    &        800                  &  \hspace{0.5cm}81.4        \\
DINO~\cite{caron2021dino}  &            &          &    800    &  3200                           &  \hspace{0.5cm}82.0$^{\dagger}$        \\
iBOT~\cite{zhou2021ibot} &            &          &    800    & 3200                              &  \hspace{0.5cm}82.3$^{\dagger}$        \\
%MAE~\cite{he2021mae}     &            &          &        &                                      &  \hspace{0.5cm}\textcolor{red}{???}           \\
MoCov3~\cite{chen2021mocov3}  &            &            &    300    &        600                 &  \hspace{0.5cm}81.4         \\
MimCo (Ours)  &                           &    MoCov3-ViT-S/16-300e        &    100    &       700      &  \hspace{0.5cm}82.5           \\ 
MimCo (Ours)  &                           &    MoCov3-ViT-S/16-300e        &    300    &          900   &  \hspace{0.5cm}\textbf{82.7}           \\ 
\midrule

Sup.~\cite{touvron21deit}   & \multirow{9}{*}{ViT-B/16} &            &        &                  &  \hspace{0.5cm}81.8         \\
BEiT~\cite{bao2021beit}  &                           &            &     800   &     800          &  \hspace{0.5cm}83.2         \\ 
DINO~\cite{caron2021dino}  &                           &            &     400   &      1600      &  \hspace{0.5cm}83.6$^{\dagger}$         \\ 
MAE~\cite{he2021mae}  &                           &            &     1600   &       1600         &  \hspace{0.5cm}83.6         \\ 
SimMIM~\cite{xie2021simmim}  &                           &            &     800   &     800      &  \hspace{0.5cm}83.8         \\ 
iBOT~\cite{zhou2021ibot}  &                           &            &     400   &         1600    &  \hspace{0.5cm}83.8$^{\dagger}$         \\ 
MoCov3~\cite{chen2021mocov3}  &                           &            &     300   &      600    &  \hspace{0.5cm}83.2         \\ 
MimCo (Ours)  &             &  MoCov3-ViT-B/16-300e          &    100    &      700                     &  \hspace{0.5cm}83.7        \\ 
MimCo (Ours)  &             &  MoCov3-ViT-B/16-300e          &    300    &     900                      &  \hspace{0.5cm}\textbf{83.9}         \\ 
\bottomrule
\end{tabular} 
}
\end{table*}

%% pseudo code
\renewcommand{\algorithmicrequire}{ \textbf{Input:}}
\renewcommand{\algorithmicensure}{ \textbf{Onput:}}

\begin{algorithm}
\caption{Pytorch-like Pseudo-code of MimCo.}
\label{alg:pseudo_code}

\begin{algorithmic}%[1]
\REQUIRE ~~\\
\textcolor{gray}{\# the learnable encoder and frozen contrastive teacher model} \\
$\mathbf{f}$,\; $\mathbf{f}^{'}$;  \\ 

\textcolor{gray}{\# the patch-level projector, momentum projector, and predictor} \\
$\mathbf{p}_1^p$,\; $\mathbf{p}_3^p$,\; $\mathbf{p}_2^p$; \\

\textcolor{gray}{\# the image-level projector, momentum projector, and predictor} \\
$\mathbf{p}_1^I$,\; $\mathbf{p}_3^I$,\; $\mathbf{p}_2^I$;\\

\STATE 
\FOR{$x$ in loader}
\STATE \textcolor{gray}{\# apply weak augmentation on images}
\STATE $x$ = augment($x$)
\STATE $m$ = random\_mask\_generator(mask\_ratio, patch\_size)
\STATE \textcolor{gray}{\# extract features for masked and non-masked images}
\STATE $z,\; z_k = \mathbf{f}(x, m),\; \mathbf{f}^{'}(x)$
\STATE $z_k = z_k.\text{detach}()$
\STATE 

\STATE \textcolor{gray}{\# extract features of masked patches: $N$ $\times$ C}
\STATE $z^p,\; z_k^p = \mathbf{p}^p_2(\mathbf{p}^p_1 (z)),\; \mathbf{p}^p_3 (z_k)$
\STATE $z^p,\; z^p_k = z^p[m],\; z^p_k[m]$
\STATE \textcolor{gray}{\# compute contrastive loss for masked patches}
\STATE $\mathcal{L}_{\text{patch}} = \text{contrastive}\_\text{loss} (z^p, z^p_k, \text{queue}\_\text{patch})$
\STATE enqueue\_dequeue(\;queue\_patch, $mean (z_k^p)$\;)

\STATE 
\STATE \textcolor{gray}{\# extract features for whole images: B $\times$ C}
\STATE $z^I,\; z^I_k = avg(z),\; avg(z_k)$
\STATE $z^I,\; z^I_k = \mathbf{p}^I_2 (\mathbf{p}^I_1 (z^I)),\; \mathbf{p}^I_3 (z^I_k)$
\STATE \textcolor{gray}{\# compute contrastive loss for images}
\STATE $\mathcal{L}_{\text{image}} = \text{contrastive}\_\text{loss} (z^I, z^I_k, \text{queue}\_\text{image})$
\STATE enqueue\_dequeue(\;queue\_image, $z^I_k$\;)

\ENDFOR

%\ENSURE ~~\\  output

%\RETURN ~~\\ return
\end{algorithmic}
\end{algorithm}

\subsection{Implementation}

\paragraph{\textbf{Architecture}}
We use the Vision Transformers~\cite{dosovitskiy2020vits} and Swin Transformers~\cite{liu2021swins} as the backbone. 
For ViTs, we conduct experiments on ViT-S and ViT-B with patch size set to 16.
For Swins, we conduct experiments on Swin-T and Swin-B with patch size set to 4 and window size set to 7.

\paragraph{\textbf{Pre-training Setup}}

We by default pre-train MimCo on ImageNet-1K training set with AdamW~\cite{LoshchilovH19adamw} optimizer and a batch size of 2048.
For ViT-S and ViT-B, we use the MoCov3~\cite{chen2021mocov3} pre-trained models as the contrastive teacher models.
For Swin-T and Swin-B, we use the MoBY~\cite{xie2021swinselfsupervised} pre-trained models as the contrastive teacher models.
If not specified, we pre-train all architectures with 100 epochs.
The learning rate is linearly warmed up during the first 10 epochs to its base value scaled with the total batch size: $\text{lr} = 1{e}^{-3} \times \text{batch}\_\text{size}~/~512$, and the weight decay is 0.05.
A light data augmentation strategy is used: random resize cropping with scale range of [0.67, 1] and a aspect ratio range of [3/4, 4/3], followed by a random flipping and a
color normalization steps.
Following SimMIM~\cite{xie2021simmim}, the default masking strategy of MimCo is: a random masking strategy with a patch size of 32$\times$32 and a mask ratio of 60\%.

\section{Experiments}

We first transfer MimCo to downstream tasks, following the standard evaluation protocols adopted in prior arts.
For the classification task on ImageNet-1K, we evaluate the quality of MimCo pre-training with Swin-T, Swin-B, ViT-S and ViT-B as backbones.
For other dense tasks, including instance detection and segmentation on MS-COCO, semantic segmentation on ADE20K, we use Swin-T as the backbone to evaluate the transfer performance of MimCo pre-training.
We then give a brief ablation study on the crucial composition of MimCo.

\subsection{Transferring Performance on Downstream Tasks}

\paragraph{\textbf{Classification on ImageNet-1K}}

%We agree with MAE~\cite{he2021mae} that, linear separability is not well correlated with transfer learning performance, especially for MIM-based pretraining.
Previous work~\cite{he2021mae,xie2021simmim} have shown that the accuracy of linear probing is not always consistent with that of finetuning, especially for MIM-based pretraining methods.
In this work, we directly study the finetuning accuracy on ImageNet-1K dataset. 
We focus on the comparison with self-supervised methods for Transformers and its supervised baseline. 
By default, we follow the finetuning protocol in iBOT~\cite{zhou2021ibot} to use a layer-wise learning rate decay, weight decay and AdamW optimizer.
Following the common practice of other self-supervised work, we search the hyperparameters for optimal fintuning performance, as shown in Table~\ref{tbl:abl_finetune_recipes}.
Expressly, for Swin-T, we set the layer-wise learning rate decay to 0.75, the drop path rate to 0.1, and the finetuning epoch to 100.
For Swin-B, we set the layer-wise learning rate decay to 0.75, the drop path rate to 0.2, and the finetuning epoch to 100.
For Vit-S/16, we set the layer-wise learning rate decay to 0.75, the drop path rate to 0.1, and the finetuning epoch to 300.
For Vit-B/16, we set the layer-wise learning rate decay to 0.65, the drop path rate to 0.1, and the finetuning epoch to 100.
%for Swin-T and Swin-B, we set the layer-wise learning rate decay to 0.75 and the finetuning epoch to 100.
%For ViT-S/16, we set the layer-wise learning rate decay to 0.75 and the finetuning epoch to 300.
%For ViT-B/16, we set the layer-wise learning rate decay to 0.65 and the finetuning epoch to 100.

As shown in Table~\ref{tbl:main_cls}, 
when pre-trained with 100 epochs, MimCo achieves top-1 accuracies of 81.7\%, 84.0\%, 82.5\%, and 83.7\% with Swin-T, Swin-B, ViT-S/16, and ViT-B/16, respectively, outperforming the contrastive teacher models and performing on par with state-of-the-art methods.
When pre-trained with 300 epochs, MimCo achieves top-1 accuracies of 81.9\%, 84.3\%, 82.7\%, and 83.9\% with Swin-T, Swin-B, ViT-S/16, and ViT-B/16, respectively,
reaching new state-of-the-art results.

%\textcolor{red}{It is also worth noting that there is a steady 0.2\%$\sim$0.3\% performance gain when increasing the finetuning epochs from 100 to 300, which demonstrates the robustness of MimCo pre-training.}

\begin{figure}[h]
\centering  
\includegraphics[width=0.9\linewidth]{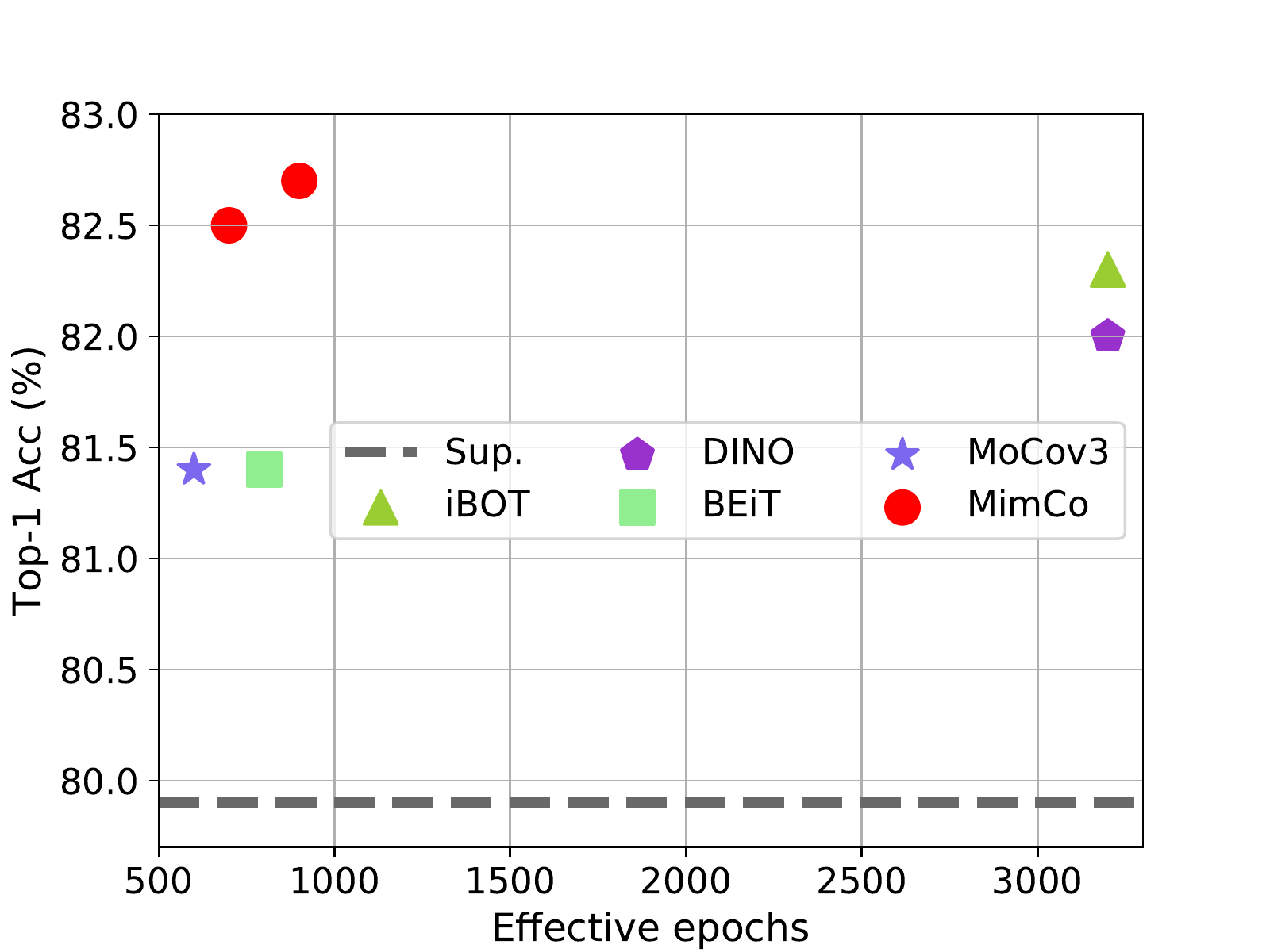}
\caption{Finetuning accuracy on ImageNet-1K \wrt~effective pre-training epochs based on ViT-S/16 architecture. Our MimCo exhibits both better transfer performance and higher pre-training efficiency.}
\Description{This figure compares the Top-1 finetuning accuracy on ImageNet-1K with regard to the effective pre-training epochs based on ViT-S/16 backbone. MimCo can achieve significantly higher finetuning accuracy using fewer pre-training epochs.}
\label{fig:trainefficiency}   
\end{figure}

Due to different training strategies, different methods with the same pre-training epochs actually see different total numbers of images.
For fair comparison of pre-training efficiency, we follow iBOT~\cite{zhou2021ibot} and use {\itshape effective pre-training epochs}, defined as actual pre-training epochs multiplied with a scaling factor accounting for extra trained images.
%
%\sout{As shown in Table~\ref{tbl:main_cls}, when pre-training with only 100 epochs, MimCo achieves top-1 accuracies of 81.64\% and 82.41\% using Swin-T and ViT-S respectively, which outperforms the previous state of the arts and takes smaller effective-pretraining-epochs.}
Taking ViT-S as the encoder, as shown in Figure~\ref{fig:trainefficiency}, our MimCo achieves a better balance between transfer performance and effective pre-training epochs compared to other pre-training methods.

%\subsection{Transfer to Downstream Tasks}
%In this section, we evaluate the effectiveness of our MimCo framework in downstream tasks, including image classification on ImageNet-1K (IN1K)~\cite{deng2009imagenet}, object detection and instance segmentation on COCO~\cite{lin2014microsoft}, and semantic segmentation on ADE20K~\cite{zhou2019semantic}.

%\subsubsection{Image Classification}
%
%Image classification is a fundamental task in computer vision, which aims to classify a given image into its corresponding class category. 

\paragraph{\textbf{Object Detection and Instance Segmentation}}

Mask R-CNN~\cite{he2017maskrcnn} is adopted in the evaluation, following the implementation of \cite{liu2021swins}.
Table~\ref{tbl:coco_det} shows a comparison of the learned representations of MimCo and other counterparts. 
MimCo pre-trained with 100 epochs achieves 43.9\% AP and 40.1\% AP on object detection and instance segmentation, respectively, outperforming sup. and MoBY~\cite{xie2021swinselfsupervised} pre-training.
When pre-trained with 300 epochs, the AP for object detection and instance segmentation are further improved to 44.9\% and 40.7\%, respectively.

\begin{table}[!h]
\caption{Results of object detection and instance segmentation finetuned 12 epochs on MS-COCO dataset. We use Mask R-CNN framework with Swin-T as the backbone. $^{*}$ denotes our training result using the official code.}
\label{tbl:coco_det}
\centering
\resizebox{0.97\linewidth}{!}{
\begin{tabular}{@{}l|c|c|c@{}}
\toprule
Method                                  & Pre-train Epochs  & $\text{mAP}^{\text{bbox}}$ (\%)   & $\text{mAP}^{\text{mask}}$ (\%) \\ \midrule
Sup.~\cite{liu2021swins}                & 100               & \hspace{0.32cm}41.6$^{\ast}$         & \hspace{0.32cm}38.4$^{\ast}$     \\ 
Sup.~\cite{liu2021swins}                & 300               & \hspace{0.2cm}43.7                & \hspace{0.2cm}39.8     \\ 
MoBY~\cite{xie2021swinselfsupervised}   & 100               & \hspace{0.2cm}41.5                & \hspace{0.2cm}38.3     \\
MoBY~\cite{xie2021swinselfsupervised}   & 300               & \hspace{0.2cm}43.6                & \hspace{0.2cm}39.6     \\ 
%RePre~\cite{wang2022repre}   & 100   & 42.1     & 39.2     \\
%RePre~\cite{wang2022repre}   & 300   & 44.8     & 40.3     \\ 
\midrule
MimCo (Ours)                            & 100               &  \hspace{0.2cm}43.9               &  \hspace{0.2cm}40.1      \\ 
MimCo (Ours)                            & 300               &  \hspace{0.2cm}\textbf{44.9}      &  \hspace{0.2cm}\textbf{40.7}      \\ \bottomrule
\end{tabular}
}
\end{table}

\paragraph{\textbf{Semantic Segmentation}}

The UPerNet~\cite{xiao2018unified} segmentation approach and the ADE20K dataset are adopted in the evaluation, following MoBY~\cite{xie2021swinselfsupervised}.
Table~\ref{tbl:main_sem_seg} shows the comparison of MimCo and other pre-training methods on this evaluation.
When pre-trained with 300 epochs, MimCo achieve an mIoU of 45.40\%, outperforming supervised and other self-supervised pre-training methods.

\begin{table}[!h]
\caption{Transfer performance comparison of ADE20K semantic segmentation. 
All models are finetuned for 160K iterations on the ADE20K dataset, with Swin-T and ViT-B/16 as the backbone and UperNet as the segmentation framework.
}
\label{tbl:main_sem_seg}
\centering
\resizebox{0.85\linewidth}{!}{
\begin{tabular}{@{}c|l|c|c@{}}
\toprule
Backbone                & Method & Pre-train Epochs & mIoU (\%)  \\ \midrule
\multirow{5}{*}{Swin-T} &   Sup.~\cite{liu2021swins}    &       &   44.51    \\ 
                        &   SimMIM~\cite{xie2021simmim} & 800   & 40.47 \\ 
                        &   MoBY~\cite{xie2021swinselfsupervised} & 300 & 44.06 \\
%\cmidrule(lr){2-4}
                        &   MimCo (Ours) & 100 & 44.44 \\
                        &   MimCo (Ours) & 300 & \textbf{45.40} \\
\midrule

\multirow{4}{*}{ViT-B/16} & Sup.~\cite{liu2021swins}    &                                        &   46.6    \\ 
                          & BEiT~\cite{bao2021beit}    &                               800      &   45.8    \\
                          & MAE~\cite{he2021mae}     &                               1600     &   48.1    \\
%\cmidrule(lr){2-4}
                          & MimCo (Ours) &                          300  &  \textbf{48.91} \\ 

\bottomrule
\end{tabular}
}
\end{table}

\begin{figure}[h]
\centering  
\includegraphics[width=0.98\linewidth]{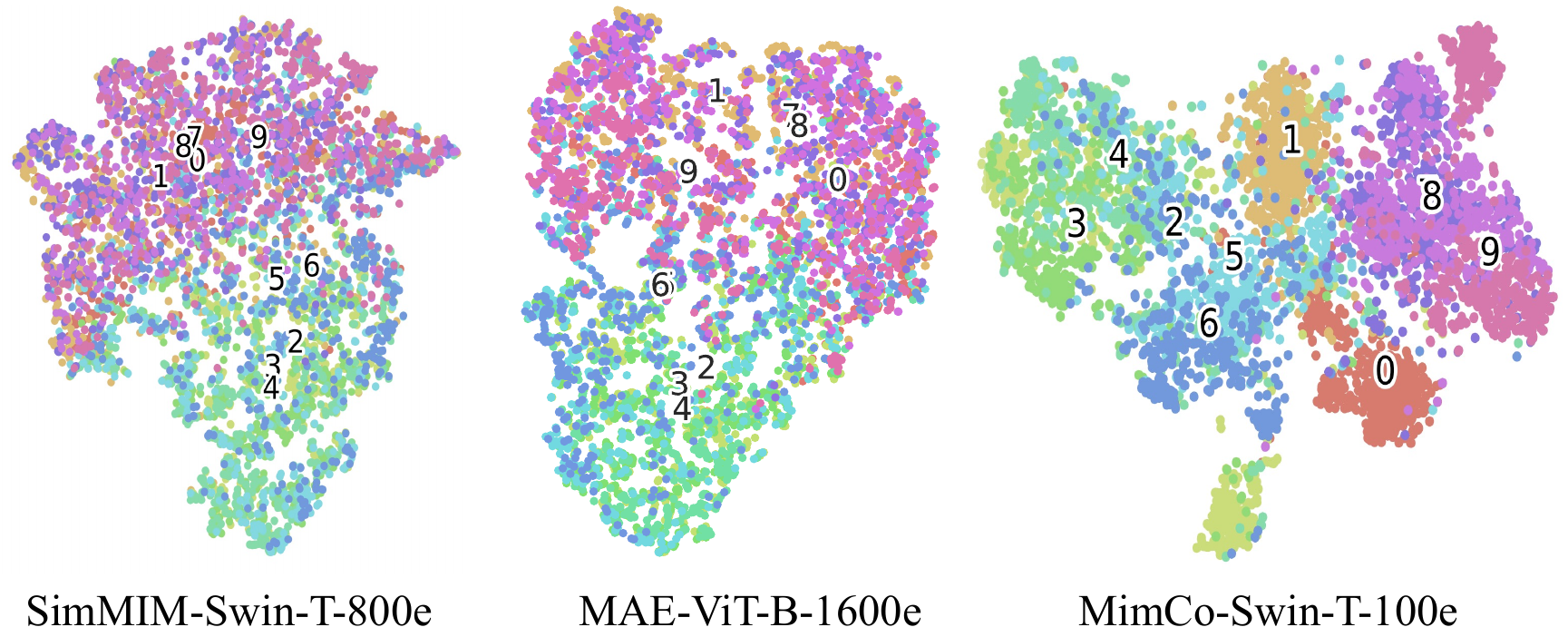}
\caption{T-SNE feature visualization of MIM method SimMIM~\cite{xie2021simmim}, MAE~\cite{he2021mae}, and our MimCo on ImageNet-1K dataset. The weights of SimMIM and MAE are from their released pre-trained models.} 
\Description{This figure shows the T-SNE feature visualization of SimMIM, MAE, and our MimCo. Compared with SimMIM and MAE, MimCo can significantly improves the linear separability of representations.}
\label{fig:motivation}   
\end{figure}

\paragraph{\textbf{Nearest Neighbor Retrieval}}
As shown in Figure~\ref{fig:motivation}, we visualize the learned features of pre-trained models using T-SNE tools. 
We randomly choose 10 classes of ImageNet-1K dataset to visualize for simplicity, 
%SimMIM~\cite{xie2021simmim} is not good at learning linear separable features.
the visualization of learned representation shows that our MimCo significantly improves the linear separability of representations compared to SimMIM~\cite{xie2021simmim} and MAE~\cite{he2021mae}.
We further evaluate MimCo on the nearest neighbor retrieval task, which is highly correlated with the linear separability of learned representations.
We consider the revisited~\cite{RadenovicITAC18} Oxford and Paris image retrieval datasets.
They contain 3 different splits of gradual difficulty with query/database pairs. We
report the Mean Average Precision (mAP) for the Medium
(M) and Hard (H) splits.
We compare MimCo with SimMIM~\cite{xie2021simmim} following the evaluation protocol as in DINO~\cite{caron2021dino}.
As reported in Table~\ref{tbl:nnr}, MimCo achieves significantly better performance on this task, further validating that the linear separability of the learned representation is improved.

\begin{table}[!h]
\caption{Effect of pre-trained features on nearest neighbor retrieval when using Swin-T as the backbone.
%The model weights of MoBY is from the official released model, and 
The model weights of SimMIM is from our pre-trained model based on the official released code.
}
\label{tbl:nnr}
\centering
\resizebox{0.97\linewidth}{!}{
\begin{tabular}{@{}l|c|cccc@{}}
\toprule
\multirow{3}{*}{Method} & \multirow{3}{*}{Pre-train Epochs} & \multicolumn{4}{c}{Image Retrieval}                                          \\ \cline{3-6} %\cmidrule(l){3-6} 
                        &                         & \multicolumn{2}{c|}{$\mathcal{R}$Ox}                        & \multicolumn{2}{c}{$\mathcal{R}$Par}   \\ \cline{3-6}  %\cmidrule(l){3-6} 
                        &                         & \multicolumn{1}{c|}{M} & \multicolumn{1}{c|}{H} & \multicolumn{1}{c|}{M} & H \\ \midrule
SimMIM~\cite{xie2021simmim} & 800    & \multicolumn{1}{c|}{4.23}  & \multicolumn{1}{c|}{1.53}  & \multicolumn{1}{c|}{8.06}  & 3.13  \\ 
%MoBY~\cite{xie2021swinselfsupervised} & 300    & \multicolumn{1}{c|}{31.18}  & \multicolumn{1}{c|}{8.51}  & \multicolumn{1}{c|}{53.25}  & 23.72  \\ 
%MimCo (Ours) & 100    & \multicolumn{1}{c|}{\textbf{32.06}}  & \multicolumn{1}{c|}{\textbf{9.00}}  & \multicolumn{1}{c|}{\textbf{53.69}}  & \textbf{24.06}  \\ 
MimCo (Ours) & 100    & \multicolumn{1}{c|}{\textbf{30.16}}  & \multicolumn{1}{c|}{\textbf{7.91}}  & \multicolumn{1}{c|}{50.82}  & 21.31  \\ 
MimCo (Ours) & 300    & \multicolumn{1}{c|}{28.73}  & \multicolumn{1}{c|}{7.81}  & \multicolumn{1}{c|}{\textbf{51.51}}  & \textbf{22.14}  \\ 
\bottomrule
\end{tabular}
}
\end{table}

\subsection{Ablation Study}

Unless otherwise specified, all ablation experiments are pre-trained for 100 epochs on ImageNet-1K dataset with Swin-T as the backbone.

%% move to supp. in final version
\begin{comment}

\paragraph{\textbf{Data Augmentation}}

Table~\ref{tbl:aug} analysis the impact of data augmentation on our pre-training framework MimCo. We inherit the simple data augmentations from MAE~\cite{he2021mae} and SimMIM~\cite{xie2021simmim}, \ie, random resized crop and random horizontal flip. 
%
Considering that we introduce contrastive learning loss into our MimCo framework, we also try some stronger augmentations, \eg, color jitter, and gaussian blur. 
The results in Table~\ref{tbl:aug} show that adding color jitter and guassian blur will degrade the performance and thus we do not use them in other experiments. 
%
In our MimCo, the random masking augmentation is strong enough for both MIM and contrastive learning.

\begin{table}[!h]
\caption{Effect of different data augmentations in our pre-training framework. ``crop'' denotes random resized crop, ``flip'' denotes random horizental flip. The experiments are performed with Swin-T as the backbone.}
\label{tbl:aug}
\centering
\resizebox{0.73\linewidth}{!}{
\begin{tabular}{l|c@{}}
\toprule
Aug. & Top-1 acc (\%) \\ \midrule
crop + flip               &        \textbf{81.66}       \\
crop + flip + color jitter    &       81.62  \\
crop + flip + gaussian blur    &      81.45  \\
\bottomrule
\end{tabular}
}
\end{table}

\end{comment}

\paragraph{\textbf{Mask Ratio}}

For pre-training, we follow the masking strategy in SimMIM~\cite{xie2021simmim} by default, which uses a patch size of 32$\times$32 and a mask ratio of 60\%. Considering that this masking strategy may not be suitable for our MimCo framework, we study how masking strategy affect the effectiveness of pre-training. 
We mainly analysis the effect of mask ratio and report the finetuning accuracy on ImageNet-1K in Table~\ref{tbl:abl_maskration}.
We empirically find that the mask ratio of 60\% performs better, and we use it for all other experiments.

\begin{table}[!h]
\caption{Effect of mask ratio in our pre-training framework. The patch size is fixed to 32$\times$32. All experiments are performed with Swin-T as the backbone.}
\label{tbl:abl_maskration}
\centering
\resizebox{0.5\linewidth}{!}{
\begin{tabular}{c|c@{}}
\toprule
Mask ratio & Top-1 acc (\%) \\ \midrule
    50\%               &        81.56        \\
     60\%               &        \textbf{81.66}        \\
    70\%               &        81.54           \\ 
\bottomrule
\end{tabular}
}
\end{table}

%% move to supp. in final version
\begin{comment}

\paragraph{\textbf{Queue Length}}

The experimental results of MoBY~\cite{xie2021swinselfsupervised} show that the pre-training performance is stable across various queue sizes after using the asymmetric encoders. Based on their experiments, we take 4096 as the default queue size in MimCo.
%
Further, we ablate the queue length in MimCo. Table~\ref{tbl:queue_size} shows that MimCo's performance is stable when the queue size varies from 2048 to 65536. However, when removing the queue design (as in MoCov3), the performance drops slightly to 81.47, probably because we use a smaller batch size of 2048.

\begin{table}[!h]
\caption{Ablation on the length of queue.}
\label{tbl:queue_size}
\centering
\small
\begin{tabular}{@{}c|c|c@{}}
\toprule
Arch.                   & queue length & Top-1 acc (\%) \\ \midrule
\multirow{7}{*}{Swin-T} & None         &     81.47           \\ 
                        & 2048         & 81.68        \\
                        & 4096         &     81.66           \\ 
                        & 8192         & 81.59        \\
                        & 16384         &     81.62      \\
                        & 32768         &   81.73        \\
                        & 65536         & 81.56     \\
                        \bottomrule
\end{tabular}
\end{table}

\end{comment}

\paragraph{\textbf{Finetuning Recipes on ImageNet-1K}}
Following the practice of previous work, we search several critical parameters (mainly the drop path rate and layer-wise learning rate decay) for the best finetuning performance.
The ablation results are reported in Table~\ref{tbl:abl_finetune_recipes}.

%We follow the finetuning protocol in SimMIM~\cite{xie2021simmim} by default, which use a layer-wise learning rate decay, a drop path rate, and a AdamW optimizer with a weight decay. 
%
%We analysis parameters sensitivity and compare the impact of different finetuning recipes to MimCo with different backbones.
%As shown in Table~\ref{tbl:abl_finetune_recipes}, we empirically find that \textcolor{red}{XXX}

\begin{table}[!h]
\caption{Different finetuning recipes on ImageNet-1K. ``L.D.'' denotes layer-wise learning rate decay, ``D.P.R.'' denotes drop path rate.}
\label{tbl:abl_finetune_recipes}
\centering
\resizebox{0.9\linewidth}{!}{
\begin{tabular}{c|c|c|c|c}
\toprule
Arch.   & Pre-train Epochs       & D.P.R. & L.D.     &  Top-1 acc (\%) \\ 
\midrule
\multirow{4}{*}{Swin-T} & \multirow{4}{*}{100} &  0.2  & 0.75     &  80.94   \\
      &  & 0.1 & 0.65 & 81.58 \\
      &  & 0.1 & 0.75 & \textbf{81.66} \\
      &  & 0.1 & 0.85 & 81.61 \\
\midrule
\multirow{5}{*}{Swin-B} & \multirow{5}{*}{100} & 0.15  & 0.75     &  83.79   \\
    & & 0.20 & 0.75 & 83.88 \\
    & & 0.25 & 0.75 & 83.80 \\
    & & 0.20 & 0.80 & \textbf{84.04} \\
    & & 0.20 & 0.85 & 83.94 \\
\midrule

\multirow{4}{*}{ViT-S} & \multirow{4}{*}{100}  & 0.2 & 0.75  & 82.28 \\
         &   & 0.1   &  0.65    &    82.34 \\
         &   & 0.1 & 0.75 & \textbf{82.53} \\
         &   & 0.1 & 0.85 & 82.49 \\
\midrule
\multirow{4}{*}{ViT-B} & \multirow{4}{*}{300} &  0.2	 & 0.65	    &  83.86  \\
                    &    &  0.1   & 0.65     &   \textbf{83.89}  \\
                    &    &  0.1	 & 0.7	    &   83.64 \\
                    &    &  0.1	 & 0.75	    &   83.65  \\
\bottomrule
\end{tabular}
}
\end{table}

\paragraph{\textbf{Reconstruction Losses}}

%We first compare different implementations of loss $L_{\text{patch}}$ and $L_{\text{image}}$ separately, and then we perform factor-by-factor ablation experiments of these two loss terms.
We first compare our patch-level reconstruction loss with existing work, and then we further experimentally verify the effectiveness of introducing additional image-level reconstruction loss.
MaskFeat~\cite{wei2021maskedfeat} verifies that reconstructing the features of the pre-trained model with $\ell_1$-loss outperforms reconstructing other targets, including RGB values and HOG features, so we directly compare with the $\ell_1$-loss feature reconstructions.
As shown in Table~\ref{tbl:loss_patch}, the accuracy of patch reconstruction using contrastive loss reaches 81.55\%, outperforming 81.35\% of reconstructing patch features with $\ell_1$-loss.

\begin{table}[!h]
\caption{Comparison of losses for reconstructing teacher model features at patch-level.
All experiments are pre-trained for 100 epochs and use Swin-T as the backbone.}
\label{tbl:loss_patch}
\centering
\resizebox{0.98\linewidth}{!}{
\begin{tabular}{@{}l|c|c@{}}
\toprule
Patch reconstruction loss & Extra model & Top-1 acc (\%) \\ \midrule
$\ell_1$ loss~\cite{wei2021maskedfeat}     &  \multirow{2}{*}{MoBY-Swin-T-300e}                    & 81.35                \\ %\\ \midrule
Contrastive loss (\textbf{ours})           &       &  \textbf{81.55}               \\ \bottomrule
\end{tabular}
}
\end{table}

To reveal the importance of additional image-level reconstruction loss $\mathcal{L}_{\text{image}}$ (defined in Equation~\ref{eq:inter_loss}), we conduct factor-by-factor experiments in this section. 
As shown in Table~\ref{tbl:ablation_loss}, loss $\mathcal{L}_{\text{patch}}$ and loss $\mathcal{L}_{\text{image}}$ achieve 81.55\% and 81.59\% top-1 accuracies, respectively, outperforming the supervised pre-training of 81.2\% and the MoBY teacher model of 81.40\%.
When using both losses, MimCo achieves the best results of 81.66\% top-1 accuracy.

\begin{table}[!h]
\caption{Ablation experiments on the patch- and image-level reconstruction loss terms of MimCo. 
Image classification results finetuned on ImageNet-1K are reported.
All experiments are pre-trained for 100 epochs and use Swin-T as the backbone.}
\label{tbl:ablation_loss}
\centering
\small
\resizebox{0.8\linewidth}{!}{
\begin{tabular}{cc|c}
\toprule
\multicolumn{2}{c|}{Reconstruction losses} & \multirow{2}{*}{ImageNet-1K Top-1 (\%)} \\ \cmidrule(r){1-2}
$\mathcal{L}_{\text{patch}}$ & $\mathcal{L}_{\text{image}}$        &                                    \\ \midrule
\checkmark         &                    & 81.55                              \\ 
                   & \checkmark         & 81.59                               \\ 
\checkmark         & \checkmark         & 81.66                               \\ \bottomrule
\end{tabular}
}
\end{table}

\begin{table}[!h]
\caption{Ablation experiments on the patch- and image-level reconstruction loss terms of MimCo. 
The results on the revisited Oxford and Paris image retrieval datasets are reported.
All experiments are pre-trained for 100 epochs and use Swin-T as the backbone.}
\label{tbl:ablation_loss_retrieval}
\centering
\small
\resizebox{0.87\linewidth}{!}{
\begin{tabular}{@{}cc|cccc@{}}
\toprule
\multicolumn{2}{l|}{Reconstruction losses}                             & \multicolumn{4}{c}{Image Retrieval}                                          \\ \midrule
\multicolumn{1}{c}{\multirow{2}{*}{$\mathcal{L}_\text{patch}$}} & \multirow{2}{*}{$\mathcal{L}_\text{image}$} & \multicolumn{2}{c|}{$\mathcal{R}$Ox}                        & \multicolumn{2}{c}{$\mathcal{R}$Par}   \\ \cmidrule(l){3-6} 
\multicolumn{1}{c}{}                        &                         & \multicolumn{1}{c|}{M} & \multicolumn{1}{c|}{H} & \multicolumn{1}{c|}{M} & H \\ 
\midrule
\multicolumn{1}{c}{\checkmark}    &       & \multicolumn{1}{c|}{22.46}  & \multicolumn{1}{c|}{5.5}  & \multicolumn{1}{c|}{39.16}  & 14.55  \\ 
\multicolumn{1}{c}{}    &  \checkmark     & \multicolumn{1}{c|}{31.58}  & \multicolumn{1}{c|}{9.04}  & \multicolumn{1}{c|}{53.26}  & 24.07  \\ 
\multicolumn{1}{c}{\checkmark}    &   \checkmark    & \multicolumn{1}{c|}{30.16} & \multicolumn{1}{c|}{7.91}  & \multicolumn{1}{c|}{50.82}  & 21.31  \\ 
\bottomrule
\end{tabular}
}
\end{table}

\paragraph{\textbf{Comparison with Multi-task Learning}}

A simple solution to combine contrastive learning and MIM is through multi-task learning.
We use ``SimMIM + MoBY'' to represent combining two pre-training methods of SimMIM~\cite{xie2021simmim} and MoBY~\cite{xie2021swinselfsupervised} through multi-task learning.
%
%Different from multi-task learning, MimCo adopts a previously contrastive learning pretrained model as a tokenizer (or teacher model). 
%Contrastive learning and MIM often have different pretraining settings, including data augmentation, optimizer hyperparameters, etc.
%The two-stage pretraining approach makes MimCo more flexible by decoupling contrastive learning and MIM.
%
As shown in Table~\ref{tbl:abl_comp_multi_task}, 
our MimCo achieves higher performance than the naive multi-task learning method under the same effective pre-training epoch.

\begin{table*}[!t]
\caption{Comparison with multi-tasking learning approach. All models take Swin-T as the backbone and are finetuned for 100 epochs on the ImageNet-1K dataset.}
\label{tbl:abl_comp_multi_task}
\centering
\resizebox{0.76\linewidth}{!}{
\begin{tabular}{l|c|c|c|c}
\toprule
Method        & Extra model  & Pre-train Epochs & Effective Epochs & Top-1 acc (\%) \\ \midrule
SimMIM + MoBY &       -           & 100    &      300 &    81.06                        \\ 
SimMIM + MoBY &       -           & 300    &    900 &       81.29                      \\ 
MimCo (Ours)  & MoBY-Swin-T-300e & 100    &   700 &          81.66                        \\ 
MimCo (Ours)  & MoBY-Swin-T-300e & 300    &     900 &   \textbf{81.86}      \\ \bottomrule
\end{tabular}
}
\end{table*}

\paragraph{\textbf{Remove Mask Operation}}

\begin{table}[!t]
    \caption{Effect of masking input in our pre-trainng framework. All experiments are performed with Swin-T as the backbone.}
    \label{tbl:abl_mask}
    \centering
    \resizebox{0.7\linewidth}{!}{
    \begin{tabular}{c|c}
    \toprule
     Masking image input   & Top-1 acc (\%) \\ \midrule
         &      81.23     \\ 
      $\checkmark$   &     \textbf{81.66}     \\ \bottomrule
    \end{tabular}
    }
\end{table}

To investigate whether MIM plays an important role in our pre-training framework, we try to remove the masking operation. 
In fact, our pre-training framework degenerates to a knowledge distillation framework when the masking operation is removed.
As shown in Table~\ref{tbl:abl_mask}, without masking input, the performance degenerates from 81.66\% to 81.23\%, indicating the critical role of masking operation in our framework.

\section{Discussion}

\begin{figure}[h]
    \centering
    \includegraphics[width=1.0\linewidth]{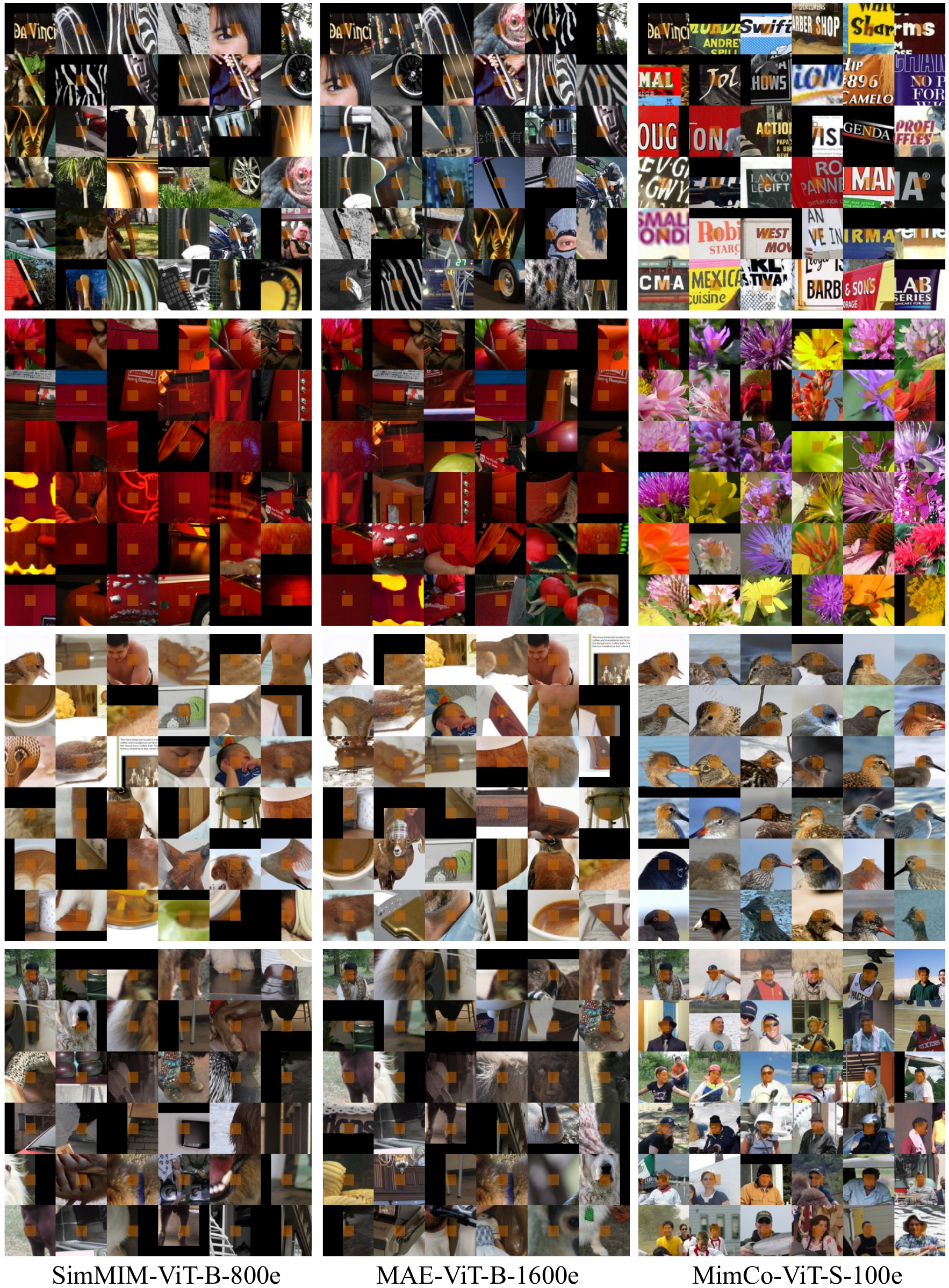}
    \caption{Visualization of semantic patterns. The top left patch is used as the query patch in each pattern layout. ``SimMIM-ViT-B-800e'' and ``MAE-ViT-B-1600e'' are from their official released pre-trained models with 800 and 1600 epochs, respectively.}
    \Description{This figure demonstrates the learned semantic patterns of SimMIM, MAE, and our MimCo. From left to right, each column denotes a pattern of ``head of person'', ``head of birds'', ``beaks'', ``colorful flowers'', and ``text on different backgrounds'', respectively.}
    \label{fig:semanticpattern}
\end{figure}

\paragraph{\textbf{What Semantic Patterns Does MimCo Learn?}}

To further help reveal what patterns does MIM learn, we follow the visualization of iBOT~\cite{zhou2021ibot} to explore the learned patterns of the pre-trained models of SimMIM~\cite{xie2021simmim}, MAE~\cite{he2021mae}, and our MimCo via visualization, respectively.
Specifically, we use the pre-trained ViT-S/16 models and visualize the top-36 most similar patches (among different images) with the highest cosine similarity on ImageNet-1K validation set. To better understand each little patch, we visualize a 80$\times$80 context for each 16$\times$16 patch (highlight in orange color).
As depicted in Figure~\ref{fig:semanticpattern}, the top left patch in each pattern layout is used as the query patch.
For all patterns, the MIM methods SimMIM~\cite{xie2021simmim} and MAE~\cite{he2021mae} tend to group the patches with similar colors regardless of their semantic meaning. This might be because they use the raw pixels as the learning target of the masked patches, which force the model to focus on learning the low-level details (\eg, {\itshape color}) and ignore high-level semantics.
%
%MoCov3 can group some patches with similar semantic meaning, \eg, {\itshape head of animals}, {\itshape creatures in sea}. However, the semantic meaning is not very consistent, \eg, it groups some patches of different animals' head, and it groups the patches of person and fish.
%
It is worth noting that, our MimCo is capable of excavating more clear and meaningful semantic patterns. \eg, {\itshape head of person}, {\itshape head of birds}, and {\itshape colorful flowers}. In addition to specific objects, the first row shows that MimCo can successfully group {\itshape text on different backgrounds}.
The results indicate that MimCo can learn both low-level details and high-level semantics at the same time.

\section{Conclusions}

%In this work we have developed a novel and flexible pre-training framework MimCo, which  

This work proposes a novel MIM pre-training framework, named MimCo, which leverages contrastive teacher models to improve the linear separability of learned representations, thereby improving pre-training performance.
MimCo is flexible and efficient: 1) the contrastive teacher model can be flexibly substituted; 2) simple weak data augmentation is used for pre-training; 3) MimCo achieves state-of-the-art transfer performance with fewer effective pre-training epochs.
We hope that our strong results and flexible pre-training framework will facilitate pre-training research, especially combining different pre-training pretext tasks such as contrastive learning and MIM.

\bibliographystyle{ACM-Reference-Format}

\balance

\bibliography{sample-base}

%\appendix


% xxx

% xxx

\end{document}